\documentclass[a4paper,conference]{IEEEtran}

\usepackage{cite}
\usepackage{booktabs}
\usepackage{graphicx}
\usepackage{amsmath}
\usepackage{amssymb}
\usepackage{xspace}
\usepackage{microtype}
\usepackage{balance}
\usepackage{placeins}
\usepackage[hidelinks]{hyperref}
\usepackage{url}

\newcommand{\systemname}{\mbox{CompressAgent}\xspace}
\newcommand{\oac}{\textsc{OAC}\xspace}
\newcommand{\glr}{\textsc{GLR}\xspace}
\newcommand{\sbc}{\textsc{SBC}\xspace}

\title{\texorpdfstring{Control Under Compression: Reliability\\Frontiers for Tool-Using Agents}{Control Under Compression: Reliability Frontiers for Tool-Using Agents}}
\author{
\IEEEauthorblockN{Yinghan Hou}
\IEEEauthorblockA{Department of Electrical and Electronic Engineering\\
Imperial College London\\
London, United Kingdom\\
\texttt{yh24@ic.ac.uk}}
\and
\IEEEauthorblockN{Zongyou Yang}
\IEEEauthorblockA{Dyson School of Design Engineering\\
Imperial College London\\
London, United Kingdom\\
\texttt{zy2926@ic.ac.uk}}
}
\date{}

\begin{document}
\maketitle

\begin{abstract}
Tool-using language-model agents are governed not only by task prompts but also by persistent system-side instructions that specify tools, arguments, policies, execution protocols, and recovery. Compressing these \emph{agent control contexts} (ACCs) can reduce input cost and context use, yet existing prompt-compression evaluations do not reveal whether the resulting control remains operationally reliable. We introduce \systemname, an environment-verified benchmark for ACC compression across nine independently constructed ACCs, three task families, three fixed Qwen API model identifiers, six retained-context budgets, and 15,525 runs. We uncover a nonlinear, method-dependent reliability frontier. At 75\% retained context, generic rewriting and section-based compression achieve 92.7\% and 92.4\% success, close to the 93.8\% full-context baseline. Between 50\% and 35\%, methods diverge sharply; at 35\%, section-based, obligation-aware, and generic rewriting achieve 47.0\%, 39.0\%, and 19.9\%. At 25--10\%, executable protocols become fragile. Reliability also varies substantially across ACCs, making universal compressor rankings inappropriate and motivating per-context qualification. Failure analysis shows that compression primarily surfaces as tool-execution and action-parsing errors. These findings recast ACC compression from token reduction into a runtime-reliability problem that must be evaluated through executable outcomes.
\end{abstract}


\section{Introduction}

Language-model agents do not operate from task text alone. Their persistent system-side context defines which tools exist, when they may be called, how arguments must be formed, which actions are prohibited, how intermediate state is maintained, and what constitutes valid termination. This context is therefore closer to a runtime control specification than to ordinary prose. Removing a seemingly minor clause can change a tool call, bypass a precondition, or make an otherwise correct trajectory unparsable.

Long prompts increase input cost and consume context capacity, motivating token pruning, rewriting, and learned compression methods~\cite{jiang2023llmlingua,pan2024llmlingua2,jiang2024longllmlingua,mu2023gist}. Yet existing compression evaluations largely emphasize question answering, reasoning accuracy, retrieval utility, or textual faithfulness. Those objectives do not directly measure whether an agent still reaches a correct environment state while obeying its control protocol. Long-context models are also sensitive to where relevant information appears~\cite{liu2024lost}, making length alone an incomplete proxy for operational reliability.

We study \emph{agent control-context compression}: budget-constrained compression of the static natural-language instructions and model-visible tool documentation supplied to an agent, while executable tools, schemas, and validators remain unchanged. \systemname evaluates nine independently constructed ACCs spanning tool use, policy compliance, and protocol-grounded multi-step execution. Each compressed artifact is frozen before held-out evaluation, and success is determined by deterministic environments rather than surface similarity.

Our central finding is a three-region \emph{reliability frontier} that turns control-context compression from a token-reduction problem into a runtime operating decision. At a 75\% budget, generic LLM rewriting (\glr) and section-based compression (\sbc) remained within 1.4 percentage points of the 93.8\% full-context success rate. In the 50--35\% transition region, methods separated sharply: at 35\%, \sbc reached 47.0\%, obligation-aware compression (\oac) 39.0\%, and \glr 19.9\%. At 25--10\%, all methods entered a high-risk region in which executable protocols became fragile. ACC-level effects varied substantially, making both budget and control-context structure central to method selection.

This paper makes four contributions:
\begin{itemize}
  \item We introduce an auditable benchmark and execution protocol for static ACC compression, covering nine ACCs, 225 held-out tasks, three fixed API model identifiers, and 15,525 logical runs.
  \item We characterize near-full, transition, and failure regions along a nonlinear and method-dependent compression--reliability frontier.
  \item We show that section structure, obligation-aware selection, and general-purpose token compression preserve executable control differently as budgets tighten.
  \item We connect artifact behavior to automatic obligation retention, feasibility warnings, tool-execution failures, and action-parsing failures.
\end{itemize}

These findings connect context engineering with agent infrastructure, AgentOps, and executable failure analysis.

\begin{figure*}[t]
  \centering
  \includegraphics[width=\textwidth]{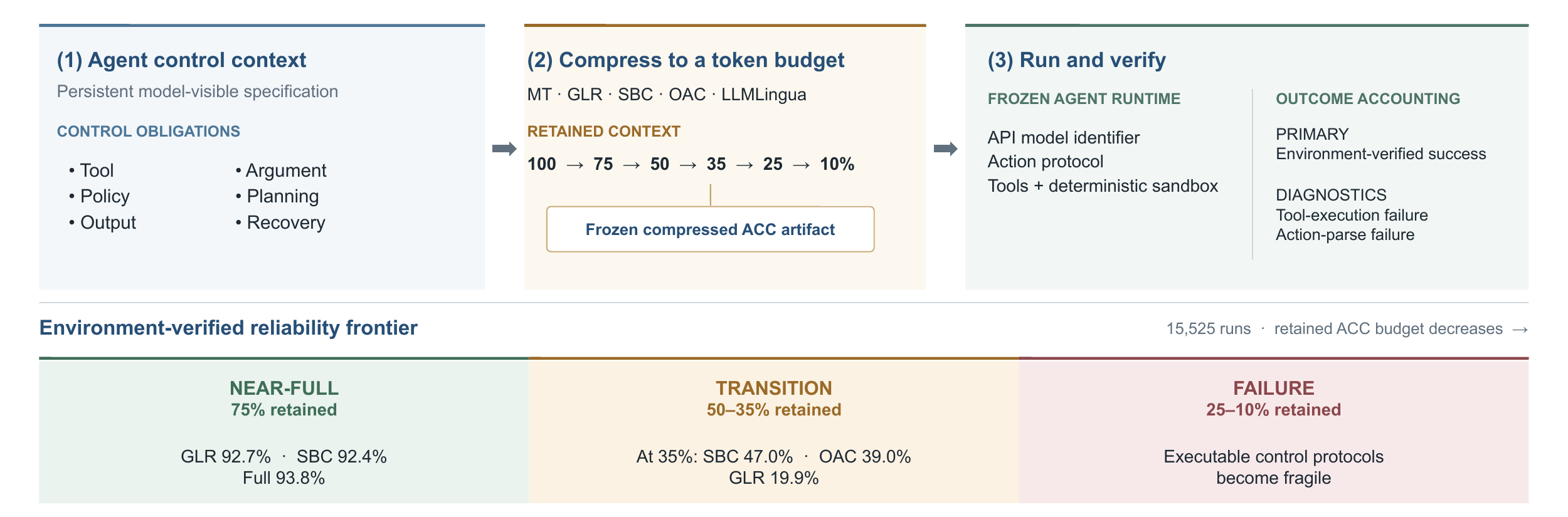}
  \caption{\systemname overview. Agent control contexts encode tool, argument, policy, planning, output, and recovery obligations. We compress only this model-visible control layer while holding executable tools and environments fixed, then evaluate frozen artifacts through environment-verified agent execution. The resulting frontier separates near-full, transition, and failure regions rather than supporting a universal compression ratio or compressor ranking.}
  \label{fig:overview}
\end{figure*}

\section{Related Work}

\paragraph{Prompt and context compression.}
LLMLingua introduced budget-controlled, coarse-to-fine prompt compression using a smaller language model~\cite{jiang2023llmlingua}; LongLLMLingua made compression query-aware for long-context tasks~\cite{jiang2024longllmlingua}; and LLMLingua-2 formulated task-agnostic compression as bidirectional token classification~\cite{pan2024llmlingua2}. Learned soft compression instead encodes prompts into compact virtual tokens~\cite{mu2023gist}. These methods establish that substantial token reduction can retain downstream utility, but their main evaluations target task content, demonstrations, or retrieved documents. \systemname instead compresses a persistent control specification and evaluates the resulting agent in an executable environment.

\paragraph{Tool-document compression.}
The closest prior work directly compresses the model-visible tool interface. Xu et al.\ retain tool and parameter names as raw tokens through selective compression and use block compression to adapt summary length to tool documentation~\cite{xu2024concise}. EASYTOOL transforms diverse, lengthy tool documentation into unified and concise tool instructions~\cite{yuan2025easytool}. Prior work optimizes tool-document representations, whereas \systemname evaluates whole control specifications---including policies, planning rules, output protocols, and recovery obligations---under matched end-to-end budgets and environment-verified execution.

\paragraph{Tool-using agents.}
ReAct interleaves reasoning with environment actions~\cite{yao2023react}; Toolformer learns when and how to invoke tools~\cite{schick2023toolformer}; and ToolLLM and Gorilla study large API collections and tool grounding~\cite{qin2024toolllm,patil2023gorilla}. In these systems, tool descriptions and protocol rules are part of the model's effective interface. We hold the tools fixed and intervene on that interface description.

\paragraph{Agent evaluation and reliability.}
AgentBench evaluates agents across interactive environments~\cite{liu2024agentbench}, AgentBoard adds progress-oriented diagnostics~\cite{ma2024agentboard}, $\tau$-bench verifies final database state under policy-guided tool use~\cite{yao2025taubench}, and AgentDojo evaluates tool agents under adversarial inputs~\cite{debenedetti2024agentdojo}. These benchmarks motivate our environment-verified outcome and failure accounting. Our complementary question is how their control layer behaves when subjected to an explicit token budget.

\section{Problem and Method}

\subsection{Agent Control Contexts}

An ACC $P_p$ is the static system-side specification presented to the model for agent $p$: natural-language instructions and model-visible tool-use documentation. Executable tool schemas, validators, and environment implementations are not compressed. As summarized in Figure~\ref{fig:overview}, we decompose $P_p$ into atomic obligations
$O_p=\{o_i\}$ of six types: tool preconditions, argument constraints, policies, planning rules, output requirements, and recovery rules. Appendix Table~\ref{tab:obligations} gives representative deletion risks.

For model $m$, task $x$, method $a$, and retained-token ratio $b$, the primary outcome is environment-verified success
\begin{equation}
Y_{m,p,x,a,b}\in\{0,1\}.
\end{equation}
The reliability frontier is the relation between $b$ and the expected value of $Y$, stratified by compression method. We report absolute success rather than normalizing away failures of the full context.

\subsection{Obligation-Aware Compression}

OAC is both a compression method and a probe of whether explicit control structure helps explain the frontier. An automatic extractor maps the full ACC to obligations with source spans, risk weights $w_i$, hard/optional status, and dependency edges. Gold obligations are reserved for external evaluation and are never available to extraction, selection, rendering, repair, or verification.

Given a token budget $B$, OAC greedily approximates the dependency-closed selection objective
\begin{equation}
\max_{\mathbf z\in\mathcal Z_{\mathrm{closed}}(E_p)}
\sum_i w_i z_i
\quad\mathrm{s.t.}\quad
L_{\mathrm{rule}}(\mathbf z) \leq B,
\end{equation}
where $\mathcal Z_{\mathrm{closed}}(E_p)$ contains selections closed under the extracted dependency graph $E_p$, and $L_{\mathrm{rule}}$ is the reference-tokenizer length of the deterministic rule rendering. Candidates are ranked deterministically by weight, length, and identifier, and admitted only when their full dependency closure remains within budget. A renderer then rewrites only selected obligations, preserving reserved identifiers and a bare-JSON action protocol. It receives at most two budget-repair attempts, after which a deterministic exact-span renderer is used.

Let $B_{\min}$ be the length of the hard-obligation dependency closure. OAC marks an artifact \emph{Feasible} when $B\geq B_{\min}$ and \emph{Best-Effort} otherwise. Best-effort runs remain in all unconditional outcomes. The alert is a pre-execution capacity check, not a predictor that the task will fail.

For mechanism analysis, automatic weighted-obligation retention is
\begin{equation}
\mathrm{WORR}=
\frac{\sum_i w_i\,\mathbb{1}[o_i\ \mathrm{selected}]}
{\sum_i w_i}.
\end{equation}
Because OAC optimizes this quantity, we use it only as an exploratory within-OAC diagnostic, not as a cross-method or causal metric.

\subsection{Research Questions and Analysis Plan}

We organize the evaluation around four questions. \textbf{RQ1} maps context budget to environment-verified reliability. \textbf{RQ2} asks whether the frontier changes by compression strategy. \textbf{RQ3} examines whether automatic obligation retention and hard-closure feasibility explain failures. \textbf{RQ4} characterizes cross-model robustness. The preregistered confirmatory analysis compares OAC with GLR at the 35\% and 25\% budgets, testing whether explicit obligation selection improves on generic semantic rewriting. All remaining method, budget, mechanism, and model analyses---including comparisons with the structural SBC baseline---are descriptive or exploratory.

\section{Benchmark and Experimental Design}

\subsection{Benchmark}

The benchmark contains three ACCs in each of three families (Table~\ref{tab:design}). Tool-use tasks cover calendar/time-zone/email, file search and summarization, and lookup/calculation/note workflows. Policy tasks cover access delegation, PII export controls, and refund escalation. Protocol-grounded tasks cover evidence citation, incident triage, and inventory restocking. Each ACC has five development tasks and 25 held-out test tasks. Test tasks are never read during artifact generation or repair.

\begin{table}[t]
\caption{Evaluation design. Full is shared across compression methods.}
\label{tab:design}
\small
\begin{tabular}{@{}p{0.35\columnwidth}p{0.60\columnwidth}@{}}
\toprule
Component & Value \\
\midrule
ACCs / task families & 9 / 3 \\
Development / test tasks & 45 / 225 \\
API model identifiers & 3 fixed Qwen endpoints \\
Budgets & 100, 75, 50, 35, 25, 10\% \\
Main compressors & MT, GLR, SBC, OAC \\
Published baseline & LLMLingua at 50, 25, 10\% target budgets \\
Main / LLMLingua slots & 14,175 / 1,350 \\
Total logical runs & 15,525 \\
\bottomrule
\end{tabular}
\end{table}

All environments are deterministic and sandboxed. The evaluator checks final environment state, valid tool and argument use, policy and protocol compliance, and termination. Multiple valid trajectories are accepted; an exact reference tool sequence is not required. No LLM-as-judge is used.

\paragraph{Task construction and quality gates.}
The nine ACCs were constructed as separate benchmark units rather than derived as paraphrases of a shared master prompt. Tasks vary goals, entities, ordering constraints, exceptional conditions, and required terminal states; they are not produced solely by swapping slot values. Before compression, every full ACC was required to complete its development suite, every environment had a legal reference trajectory, and evaluator tests covered both valid alternatives and invalid near-misses. Development tasks could be used to qualify infrastructure and full-context behavior, but not to tune an artifact after seeing held-out outcomes.

\paragraph{Outcome contract.}
A run succeeds only if the evaluator confirms the required terminal state and all applicable control conditions. A syntactically valid tool call with a wrong semantic argument is a failure, as is a correct final answer reached through a forbidden side effect. The common evaluator contract also records tool selection, argument semantics, policy and protocol compliance, recovery, termination reason, and failure modes. The paper uses binary environment success as the primary outcome because it is consistent across families and does not assign arbitrary weights to submetrics.

\subsection{Compression Methods}

\textbf{Mechanical truncation (MT)} retains a tokenizer-bounded prefix. \textbf{Generic LLM rewriting (GLR)} asks a fixed compressor to preserve tools, preconditions, policies, order, recovery, and the JSON action envelope without access to obligations or tasks; up to two uniform budget repairs and a semantic-boundary fallback enforce the budget. \textbf{Section-based compression (SBC)} deterministically selects whole semantic units using a preregistered weighted round-robin over action protocol, tool, policy, planning/recovery, and lower-priority sections while preserving source order. \textbf{OAC} uses the procedure above. All four methods are evaluated at five compressed budgets. GLR generation and OAC's LLM-backed renderer use the same frozen \nolinkurl{gemma-4-12b-it} compressor (temperature 0; seed 20260722); OAC's automatic obligation extractor is deterministic rule-based code, not an LLM. Full generation code, the model SHA-256, and per-ACC reserved-identifier overlays are included in the supplement.

We include \textbf{LLMLingua}~\cite{jiang2023llmlingua} as a published baseline on two preregistered evaluation endpoints at 50\%, 25\%, and 10\% target budgets. We use \texttt{llmlingua==0.2.2}, a fixed GPT-2 checkpoint, CPU execution, one candidate, and no manual repair. This subset contributes 1,350 logical runs.

\paragraph{Artifact generation and isolation.}
The four main methods produce $9\times5=45$ artifacts each; LLMLingua produces 27, for 207 logical artifacts in total. Compressors may read the full ACC, target budget, and their preregistered configuration. GLR is additionally forbidden from reading automatic obligations, while OAC may read only automatic obligations. All methods are forbidden from reading test tasks, gold obligations, evaluator feedback, or test failures. We generate one artifact per ACC--method--budget cell and never select the best of multiple candidates using behavior.

\subsection{Models, Runtime, and Analysis}

The three fixed Alibaba Cloud Model Studio identifiers are
\nolinkurl{qwen3.6-flash-2026-04-16},
\nolinkurl{qwen3.6-35b-a3b}, and
\nolinkurl{qwen3.7-plus-2026-05-26}.
They share an OpenAI-compatible transport, disabled thinking and search, no native function calling, temperature 0, top-$p$ 1, at most ten agent steps, and a fixed JSON action protocol. These endpoints differ in more than parameter count; model-level results are therefore robustness checks, not a scale-law analysis.

The preregistered confirmatory contrasts compare OAC with GLR at 35\% and 25\%, paired on $(m,p,x,b)$. Each ACC receives equal weight. Two-sided 95\% intervals use a 10,000-iteration hierarchical paired bootstrap that first resamples ACCs, then tasks within ACCs, while retaining the three paired model observations~\cite{efron1993bootstrap}. One-sided superiority $p$-values use an exact sign-flip test over nine ACC effects; the two co-primary tests use Holm correction~\cite{holm1979}. All other analyses are secondary or exploratory.

\paragraph{Logical runs and failure accounting.}
The shared Full baseline contributes 675 logical runs, the four main methods contribute 13,500, and LLMLingua adds 1,350, for 15,525 total. All preregistered logical runs reached terminal scientific outcomes. Infrastructure retries followed the frozen failure policy and did not alter scientific outcome definitions; complete attempt-level accounting is provided in the supplementary material. Model grammar violations, environment failures, and method-level operational incompatibility remain unconditional scientific failures, and OAC best-effort runs remain included.

\section{Results}

\subsection{A Nonlinear Reliability Frontier}

\begin{figure*}[t]
  \centering
  \includegraphics[width=\textwidth]{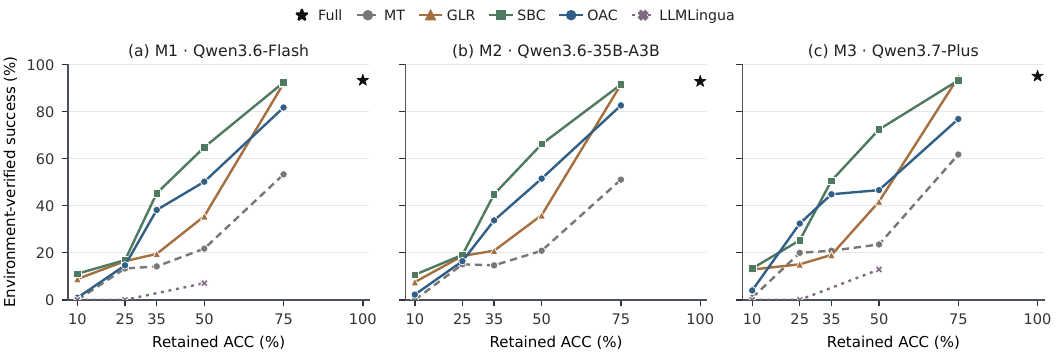}
  \caption{Environment-verified success across budgets for the three fixed API model identifiers. Full is the shared 100\% baseline; LLMLingua is evaluated only at 50\%, 25\%, and 10\% target budgets on two preregistered endpoints. The frontier is nonlinear and method-dependent.}
  \label{fig:frontier}
\end{figure*}

Figure~\ref{fig:frontier} shows a sharp, method-dependent frontier. Full ACCs achieved 633/675 successes (93.8\%). At 75\%, GLR achieved 92.7\% and SBC 92.4\%, within 1.4 points of Full descriptively. This is a near-full region in our setting, not a formal non-inferiority guarantee. MT (55.4\%) and OAC (80.4\%) demonstrate that a large token budget alone does not ensure preservation.

Reliability fell steeply between 75\% and 35\%. At 50\%, SBC led the main methods at 67.9\%, followed by OAC at 49.5\%, GLR at 37.8\%, and MT at 22.1\%. At 35\%, SBC remained highest (47.0\%), followed by OAC (39.0\%), GLR (19.9\%), and MT (16.6\%). At 25\%, all four methods clustered between 16.1\% and 21.2\%; at 10\%, none exceeded SBC's 11.7\%. Thus, no method dominates across budgets, and a compression ratio cannot be transferred independently of its selection rule.

The transition is visible for all three endpoints rather than being driven by one provider configuration. At 75\%, GLR ranged from 91.6\% to 94.7\% and SBC from 91.6\% to 93.3\%; at 35\%, GLR fell to 19.1--20.9\% and SBC to 44.9--50.7\%. OAC was more variable at 35\% (33.8--44.9\%). These are descriptive robustness ranges: endpoint identity is paired in the main analysis but not treated as an independent population.

The near-full region also yields measurable input-token and estimated API-cost savings in the evaluated runtime (Table~\ref{tab:operating-point}). At the 75\% target, GLR and SBC reduced attempt-inclusive input tokens by 21.6\% and 19.3\%, and frozen-standard-price API cost by 19.2\% and 17.0\%, respectively. Active execution latency did not improve (median 5.03\,s for GLR and 4.88\,s for SBC versus 4.21\,s for Full), so we make no latency-reduction claim.

\begin{table}[t]
\caption{Descriptive 75\% operating point. ACC tokens use the frozen reference tokenizer; input and estimated CNY cost include recorded physical retries and provider usage.}
\label{tab:operating-point}
\scriptsize
\begin{tabular}{lrrrr}
\toprule
Condition & ACC tok. & Input/run & CNY/run & Success \\
\midrule
Full & 1450.6 & 9079.4 & .01657 & 93.8\% \\
GLR 75\% & 1039.0 & 7120.8 & .01339 & 92.7\% \\
SBC 75\% & 1082.4 & 7327.4 & .01375 & 92.4\% \\
\bottomrule
\end{tabular}
\end{table}

\subsection{Preregistered Comparison Reveals Context-Specific Regimes}

At 35\%, OAC reached 39.0\% versus GLR's 19.9\%, yielding a practically substantial paired estimate of $+19.11$ percentage points (Table~\ref{tab:primary}). The ACC-level interval crossed zero, reflecting pronounced context heterogeneity rather than a uniform effect (95\% CI $[-0.59,40.59]$; Holm-adjusted $p=.1563$). At 25\%, the paired estimate was $+4.44$ points with wider uncertainty (95\% CI $[-21.19,26.08]$; Holm-adjusted $p=.3906$). The corrected tests do not support a universal OAC advantage; together with the wide ACC-level variation, they motivate the context-specific analysis below.

\begin{figure}[t]
  \centering
  \includegraphics[width=\columnwidth]{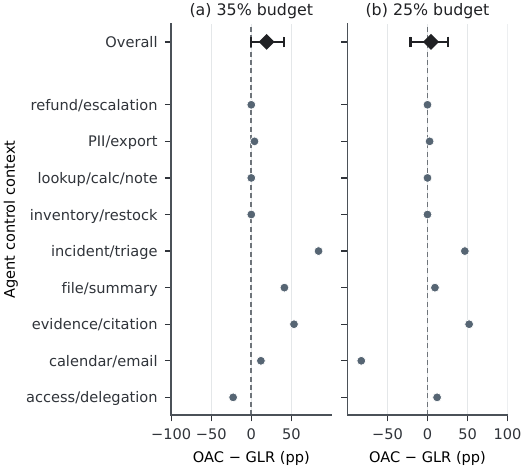}
  \caption{ACC-level effects reveal distinct compression regimes: OAC strongly benefits several control contexts, while others favor generic rewriting, motivating context-aware method selection. Positive values favor OAC; negative values favor GLR. The red interval is the equal-ACC aggregate.}
  \label{fig:paired}
\end{figure}

Figure~\ref{fig:paired} reveals the underlying regimes. At 35\%, ACC-level effects ranged from $-22.7$ points for access delegation to $+84.0$ for incident triage; three ACCs had zero success under either method. At 25\%, calendar/time-zone/email favored GLR by 82.7 points, while evidence citation and incident triage favored OAC by 52.0 and 46.7 points. Cross-model descriptive effects were consistently positive at 35\% ($+12.9$ to $+25.8$ points) but mixed at 25\% ($-2.2$ to $+17.3$). The preregistered evidence therefore motivates budget- and context-specific compressor selection rather than a universal ranking.

\begin{table}[t]
\caption{Preregistered paired OAC--GLR contrasts. CIs are two-sided; $p_H$ is the Holm-adjusted one-sided exact sign-flip $p$-value.}
\label{tab:primary}
\small
\begin{tabular}{rrrrr}
\toprule
$b$ & OAC & GLR & $\Delta$ pp [95\% CI] & $p_H$ \\
\midrule
35\% & 39.0 & 19.9 & 19.11 [$-$0.59, 40.59] & .1563 \\
25\% & 21.2 & 16.7 & 4.44 [$-$21.19, 26.08] & .3906 \\
\bottomrule
\end{tabular}
\end{table}

\subsection{The Evaluated LLMLingua Configuration Transfers Poorly to ACCs}

The evaluated LLMLingua configuration achieved 45/450 successes (10.0\%) at 50\%, and 0/450 at both 25\% and 10\%. At 25\%, every run failed action parsing; at 10\%, 449/450 did. This result is specific to the tested version, checkpoint, and unmodified artifact contract, but it exposes a transfer gap: preserving broadly informative tokens is not sufficient to preserve tool identifiers, control envelopes, and executable protocol semantics.

The contrast with the other methods is especially informative at 50\%. LLMLingua retained a mean of 607 system tokens, compared with OAC's 665, GLR's 716, and SBC's 721, yet achieved 10.0\% versus 49.5\%, 37.8\%, and 67.9\%. Token count explains only part of the gap. Nearly half of LLMLingua runs at this budget failed parsing and another 41.1\% failed during tool execution, suggesting that the remaining tokens did not form a stable control interface.

\begin{figure*}[t]
  \centering
  \includegraphics[width=\textwidth]{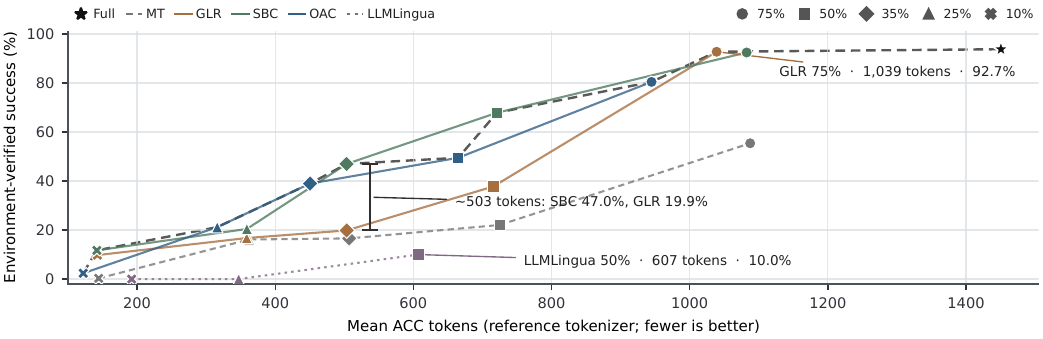}
  \caption{Control reliability at matched context lengths. Each point is one frozen method--budget condition aggregated across three API model identifiers (LLMLingua: two); marker shape denotes target budget. The dashed curve is the descriptive empirical non-dominated envelope under fewer ACC tokens and higher success, not a fitted frontier or confidence bound. Similar-length conditions can differ sharply, showing that retained token count alone does not determine executable reliability.}
  \label{fig:token-pareto}
\end{figure*}

Figure~\ref{fig:token-pareto} recasts the frontier in actual ACC length. At nearly identical 35\% lengths, SBC and GLR both used about 503 tokens but achieved 47.0\% and 19.9\% success. The LLMLingua 50\% condition likewise reached only 10.0\% at 607 tokens, despite lying near the 50\% OAC and SBC artifacts in length. The resulting spread makes token count an insufficient proxy for preserved control.

\subsection{Obligation Retention and Execution Failures}

\begin{figure*}[t]
  \centering
  \includegraphics[width=\textwidth]{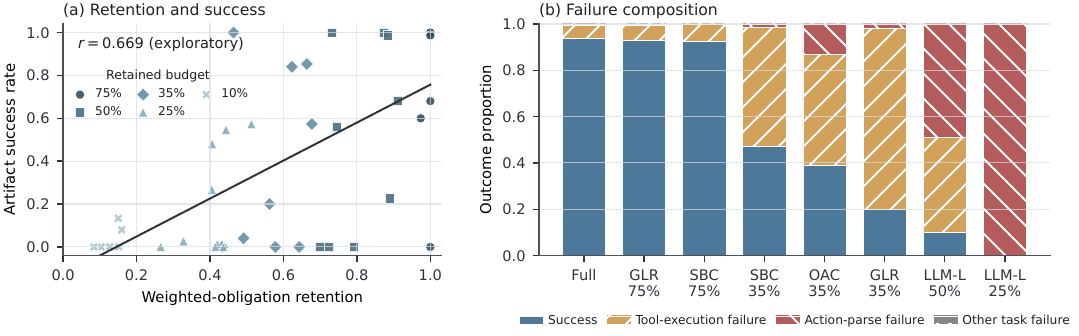}
  \caption{Mechanism and failure evidence. (a) Exploratory relation between automatic OAC weighted-obligation retention and artifact success across 45 artifacts; the line is descriptive. (b) Outcome composition for representative conditions. Tight compression manifests primarily as tool-execution or action-parsing failure.}
  \label{fig:mechanisms}
\end{figure*}

Within OAC's 45 artifacts, automatic weighted-obligation retention had an exploratory Pearson correlation of $r=0.669$ with artifact success (Figure~\ref{fig:mechanisms}a). This association is consistent with the view that \emph{what} survives matters in addition to token count, while neither establishing causality nor measuring gold semantic retention.

Feasibility warnings became common as budgets tightened. All 75\% OAC runs were \emph{Feasible}; at 50\%, 450/675 were \emph{Best-Effort}; at 35\%, 525/675; at 25\%, 600/675; and at 10\%, all 675. At 50\%, feasible artifacts achieved 85.3\% success versus 31.6\% for best-effort artifacts. Because feasibility is defined mechanically by $B<B_{\min}$, its perfect agreement with that rule is not task-failure prediction.

Across the complete ledger, 9,992 runs failed: 7,969 (79.7\%) through tool execution, 2,002 (20.0\%) through output parsing, and only 21 (0.2\%) as other task failures. The composition changes by method and budget (Figure~\ref{fig:mechanisms}b): 35\% GLR failures are overwhelmingly tool-execution failures, whereas very tight OAC and LLMLingua conditions increasingly lose the action envelope itself. Compression therefore changes an agent's operational failure mode, not merely answer accuracy.

The best-effort flag is informative but not a safety certificate. At 35\%, feasible OAC artifacts succeeded at 52.7\% and best-effort artifacts at 35.0\%; at 25\%, the corresponding rates were 26.7\% and 20.5\%. Some best-effort contexts still perform well---notably calendar/time-zone/email at 35\%---while some feasible contexts remain difficult. $B_{\min}$ answers whether the extracted hard closure fits, not whether extraction is complete or whether a model can execute the retained wording.

\section{Discussion}

\paragraph{Moderate compression preserves near-full reliability.}
At 75\%, GLR and SBC remained within 1.4 points of Full, showing that meaningful redundancy can be removed with little observed reliability loss in this setting. A practical qualification workflow can begin at this moderate budget, evaluate the artifact on executable tasks, and treat every tighter budget as a distinct runtime configuration. The observed region is an operating point, not a universal safety guarantee.

\paragraph{Section structure is a strong control-preservation primitive.}
SBC was strongest at both 50\% and 35\%. Stable section boundaries and source order preserve interactions among tool definitions, policies, protocols, and recovery clauses that token-level importance alone may fragment. This makes structural units a strong baseline and a useful design primitive for future ACC compressors.

\paragraph{Obligation signals enable risk-aware compression.}
OAC adds a capability absent from token- and section-based approaches: it exposes which control obligations are retained, whether their dependency closure fits, and when compression has entered a best-effort regime. Its automatic retention signal tracks artifact success, while $B_{\min}$ can warn before execution that the extracted hard closure no longer fits. These diagnostics do not certify safety, but they turn compression from blind shortening into an auditable capacity decision.

\paragraph{ACC compression requires per-context qualification.}
The three operational regions are deliberately context-dependent. In the \emph{near-full region} around 75\%, GLR and SBC remove redundancy with little observed reliability loss; in the \emph{transition region} around 50--35\%, method choice and ACC structure dominate; and in the \emph{failure region} at 25--10\%, executable protocols become fragile. The dominant tool-execution and parser failures further show why semantic similarity is insufficient. Deployment should locate the boundary per ACC and runtime, monitor action-envelope validity and environment outcomes, and avoid copying a global compression percentage or universal method ranking.

\paragraph{Safety and ethical scope.}
All tool calls execute in deterministic simulations: no real email is sent, no live calendar or file is modified, and no private user data is accessed. The failure modes nevertheless have deployment implications because compressing policies and recovery rules can silently weaken safeguards. ACC compression should not be used in safety-critical settings without environment-level validation and continuous monitoring. Released logs must exclude credentials, personal data, and provider secrets; our artifacts contain only synthetic tasks and redacted request metadata.

\subsection{Limitations}

Our conclusions concern nine English ACCs, three deterministic task families, and three fixed API model identifiers. Confirmatory inference is conducted at the ACC level; with nine independent ACCs, uncertainty about a universal method ranking is necessarily dominated by between-context heterogeneity, and non-significance does not establish equivalence. Model-specific results serve as robustness evidence rather than scale-law claims. The evaluated artifacts are single realizations, and WORR is an automatic OAC-internal diagnostic rather than gold cross-method semantic retention. Live tools, provider-native function calling, and dynamic schemas may shift the empirical frontier.

\FloatBarrier

\section{Conclusion}

\systemname establishes agent control-context compression as a runtime-reliability problem. Across 15,525 auditable runs, we identify a near-full region in which moderate rewriting or section-based compression preserves behavior, a transition region in which method and context structure determine reliability, and a failure region in which executable protocols become fragile. Section structure provides a strong preservation primitive, while obligation-aware feasibility and retention signals expose when critical control information no longer fits. The evaluated general-purpose token-compression baseline did not reliably preserve the tested executable interface. Reliable ACC compression therefore requires environment-level qualification of what survives and how agents fail---not token reduction alone.

\label{lastmainpage}

\FloatBarrier
\balance
\bibliographystyle{IEEEtran}
\bibliography{references}

\appendix

\section{Additional Results}

\begin{table*}[!t]
\caption{Benchmark characterization. Token counts use the frozen reference tokenizer; tools/actions exclude the terminal \texttt{finish} action.}
\label{tab:benchmark-characterization}
\centering
\small
\begin{tabular}{llrrrr}
\toprule
ACC & Family & Full tok. & Tools/actions & Gold obligations & Hard \\
\midrule
Calendar/time-zone/email & Tool use & 1348 & 3 & 18 & 16 \\
File search/summary & Tool use & 1459 & 4 & 23 & 21 \\
Lookup/calculator/note & Tool use & 1634 & 3 & 19 & 19 \\
Access delegation & Policy & 1890 & 5 & 26 & 25 \\
PII export control & Policy & 1230 & 3 & 15 & 15 \\
Refund escalation & Policy & 1398 & 4 & 16 & 16 \\
Evidence citation & Protocol & 1187 & 3 & 16 & 16 \\
Incident triage & Protocol & 1288 & 5 & 18 & 18 \\
Inventory restock & Protocol & 1621 & 5 & 18 & 18 \\
\bottomrule
\end{tabular}
\end{table*}

\begin{table}[!htbp]
\caption{ACC obligation types and representative failure risks.}
\label{tab:obligations}
\small
\begin{tabular}{lp{0.66\columnwidth}}
\toprule
Type & Control information and deletion risk \\
\midrule
Tool & Applicability and preconditions; wrong tool or activation \\
Argument & Types, formats, ranges; invalid or semantically wrong call \\
Policy & Required/forbidden behavior; safety or control violation \\
Planning & Step order and state; skipped, repeated, or looping action \\
Output & Action envelope and final schema; parser rejection \\
Recovery & Missing data and tool failure; premature termination \\
\bottomrule
\end{tabular}
\end{table}

\begin{table}[!htbp]
\caption{Cross-model aggregate frontier (success rate, \%). LLMLingua uses two endpoints.}
\small
\begin{tabular}{lrrrrrr}
\toprule
Method & 100 & 75 & 50 & 35 & 25 & 10 \\
\midrule
Full & 93.8 & -- & -- & -- & -- & -- \\
GLR & -- & 92.7 & 37.8 & 19.9 & 16.7 & 9.8 \\
MT & -- & 55.4 & 22.1 & 16.6 & 16.1 & 0.3 \\
OAC & -- & 80.4 & 49.5 & 39.0 & 21.2 & 2.4 \\
SBC & -- & 92.4 & 67.9 & 47.0 & 20.4 & 11.7 \\
LLMLingua & -- & -- & 10.0 & -- & 0.0 & 0.0 \\
\bottomrule
\end{tabular}
\end{table}

\begin{table}[!htbp]
\caption{LLMLingua artifact-contract diagnostics by target budget. ``ID complete'' requires every preregistered reserved identifier to survive.}
\label{tab:llmlingua-contract}
\small
\begin{tabular}{lrrrr}
\toprule
Target & Within 5\% & Above & Mean tok. & ID complete \\
\midrule
50\% & 9/9 & 0/9 & 607.4 & 0/9 \\
25\% & 7/9 & 2/9 & 347.1 & 0/9 \\
10\% & 3/9 & 6/9 & 192.2 & 0/9 \\
\bottomrule
\end{tabular}
\end{table}

\FloatBarrier

All 27 single-pass LLMLingua outputs were structurally loadable and retained without repair. Nineteen met the preregistered 5\% upper tolerance; all eight exceptions exceeded rather than undershot the target. Reporting actual token counts therefore gives this baseline at least its generated context, while its low execution success cannot be attributed to receiving fewer tokens than the target.

\begin{table*}[t]
\caption{API model identity and endpoint qualification. Each requested identifier matched the returned identifier in four of four development-only probes.}
\label{tab:model-identity}
\centering
\small
\begin{tabular}{llrr}
\toprule
Role & Requested and returned model identifier & Context limit & Exact probes \\
\midrule
M1 & \texttt{qwen3.6-flash-2026-04-16} & 1M & 4/4 \\
M2 & \texttt{qwen3.6-35b-a3b} & 256K & 4/4 \\
M3 & \texttt{qwen3.7-plus-2026-05-26} & 1M & 4/4 \\
\bottomrule
\end{tabular}
\end{table*}

The endpoints were qualified on 23 July 2026 with common fingerprint
\texttt{28686a0e484f\ldots735373c}; the supplementary manifest records the full value.
All runs used temperature 0, top-$p$ 1, thinking and search disabled, no native tool calling, a 1024-token completion cap, and a ten-action limit. M2 is a provider model identifier rather than a dated snapshot; every formal record therefore preserves both requested and returned identities.

\begin{table}[h]
\caption{OAC feasibility and unconditional success.}
\small
\begin{tabular}{llrr}
\toprule
Budget & State & Slots & Success (\%) \\
\midrule
75\% & Feasible & 675 & 80.4 \\
50\% & Feasible & 225 & 85.3 \\
50\% & Best-Effort & 450 & 31.6 \\
35\% & Feasible & 150 & 52.7 \\
35\% & Best-Effort & 525 & 35.0 \\
25\% & Feasible & 75 & 26.7 \\
25\% & Best-Effort & 600 & 20.5 \\
10\% & Best-Effort & 675 & 2.4 \\
\bottomrule
\end{tabular}
\end{table}
\FloatBarrier

\section{Evaluation Contract and Information Boundaries}

\paragraph{Deterministic adjudication.}
Task success is computed by deterministic code over the action trace and terminal environment state. The evaluated agent models never score their own outputs, and no separate language model assigns correctness labels. The unified evaluator contract records binary task success together with tool-selection, argument, policy, protocol, recovery, termination, and failure-mode fields. These diagnostic fields do not replace the environment-verified primary outcome.

\begin{table*}[!htbp]
\caption{Frozen artifact-generation and information-isolation contract. ``Tasks'' includes development and held-out tasks; ``feedback'' includes evaluator outputs and observed behavior.}
\label{tab:generation-contract}
\centering
\footnotesize
\setlength{\tabcolsep}{3pt}
\begin{tabular}{lp{0.29\textwidth}p{0.27\textwidth}p{0.27\textwidth}}
\toprule
Method & Frozen construction & Permitted inputs & Explicitly unavailable \\
\midrule
MT & Tokenizer-bounded prefix & Full ACC, target budget, reference tokenizer & Automatic/gold obligations, tasks, feedback \\
SBC & Deterministic whole-unit selection with frozen section priorities and source order & Full ACC, target budget, tokenizer, fixed section priorities & Automatic/gold obligations, tasks, feedback \\
GLR & Shared \texttt{gemma-4-12b-it} compressor at temperature 0; at most two budget repairs; deterministic semantic-boundary fallback & Full ACC, target budget, reference tokenizer & Automatic/gold obligations, tasks, feedback \\
OAC & Rule-based obligation extraction, dependency-closed selection, shared Gemma renderer, and traceable deterministic fallback & Full ACC, target budget, automatic obligations, reserved identifiers, tokenizer & Gold obligations, tasks, feedback \\
LLMLingua & \texttt{llmlingua==0.2.2}, frozen GPT-2 checkpoint, one candidate, no manual repair & Full ACC and target budget & Automatic/gold obligations, tasks, feedback \\
\bottomrule
\end{tabular}
\end{table*}

One artifact was generated for each ACC--method--budget cell and frozen before held-out evaluation. No artifact was regenerated, repaired, or selected according to task performance. GLR and OAC used the same compressor model with temperature-zero decoding; OAC's extractor itself was deterministic rule-based code.

\begin{table*}[!htbp]
\caption{Frozen scientific-outcome and infrastructure-accounting contract. Infrastructure events never enter the success denominator unless they resolve to a terminal scientific outcome.}
\label{tab:failure-contract}
\centering
\footnotesize
\setlength{\tabcolsep}{3pt}
\begin{tabular}{p{0.24\textwidth}p{0.20\textwidth}p{0.06\textwidth}p{0.13\textwidth}p{0.24\textwidth}}
\toprule
Observed condition & Ledger class & $Y$ & Included? & Interpretation \\
\midrule
Environment verifies all task conditions & \url{SUCCESS} & 1 & Yes & Scientific success \\
Completed run violates a task condition & \url{NORMAL_TASK_FAILURE} & 0 & Yes & Scientific failure \\
Model emits an invalid action envelope & \url{MODEL_OUTPUT_PARSE_FAILURE} & 0 & Yes & Model/protocol failure, not parser infrastructure \\
Expected tool error is not recovered & \url{TOOL_EXECUTION_TASK_FAILURE} & 0 & Yes & Scientific failure in the simulated environment \\
Compressor artifact is intrinsically invalid or operationally incompatible & \url{METHOD_ARTIFACT_INVALID} or LLMLingua incompatibility & 0 & Yes & Method-level unconditional failure; no human repair \\
API transport, harness parser, harness, or tool runtime crashes unexpectedly & Infrastructure class & -- & No & Retried at most twice; unresolved events block analysis \\
Frozen artifact is missing or corrupted after sealing & \url{ARTIFACT_STORAGE_FAILURE} & -- & No & Infrastructure failure; never converted to task failure \\
\bottomrule
\end{tabular}
\end{table*}
\FloatBarrier

\section{Confirmatory Analysis and Claim Scope}

For budget \(b\), the paired effect within ACC \(p\) is
\[
\begin{aligned}
d_{m,p,x,b} &=
Y^{\mathrm{OAC}}_{m,p,x,b}-Y^{\mathrm{GLR}}_{m,p,x,b},\\
\widehat{\Delta}_{p,b}
&= \frac{1}{75}\sum_{m=1}^{3}\sum_{x=1}^{25}d_{m,p,x,b},\\
\widehat{\Delta}_{b}
&= \frac{1}{9}\sum_{p=1}^{9}\widehat{\Delta}_{p,b}.
\end{aligned}
\]
Thus, every ACC receives equal top-level weight, while model and task observations remain paired within ACC. The 10,000-iteration interval resamples ACCs first and tasks within resampled ACCs second, retaining all three paired model observations. The one-sided exact sign-flip test operates on the nine ACC effects; Holm correction covers the two preregistered budgets.

\begin{table*}[!htbp]
\caption{Analysis status, evidence, and intended scope. This table distinguishes preregistered inference from descriptive and exploratory evidence.}
\label{tab:claim-scope}
\centering
\footnotesize
\setlength{\tabcolsep}{4pt}
\begin{tabular}{p{0.18\textwidth}p{0.12\textwidth}p{0.23\textwidth}p{0.35\textwidth}}
\toprule
Analysis & Status & Evidence & Scope boundary \\
\midrule
OAC--GLR at 35\% and 25\% & Confirmatory & Paired risk differences, ACC-first intervals, exact tests, Holm correction & Tests superiority at the two frozen budgets; it does not establish equivalence or a universal compressor ranking \\
Reliability frontier across methods and budgets & Descriptive & Full budget grid, model robustness ranges, token--success envelope & Identifies empirical operating regions in this benchmark; it is not a fitted change point or non-inferiority guarantee \\
OAC WORR--success association & Exploratory & Pearson \(r=0.669\) across 45 OAC artifacts & OAC-internal automatic diagnostic; neither causal nor cross-method gold semantic retention \\
Failure composition & Descriptive & Terminal ledger classes and representative outcome compositions & Characterizes how failures surface; it does not identify a causal missing obligation \\
LLMLingua transfer & Secondary & 27 frozen artifacts on two preregistered endpoints & Descriptive result for the evaluated package, checkpoint, and artifact contract; not every general-purpose compressor \\
\bottomrule
\end{tabular}
\end{table*}
\FloatBarrier

\section{Reproducibility and Audit Boundaries}

The release binds the design, benchmark, model-runtime amendments, 207 compressed artifacts, formal run ledger, primary analysis, secondary analysis, and post-execution audit with SHA-256 manifests. A logical run identifier is stable across infrastructure retries; exactly one attempt becomes its official scientific outcome. Infrastructure failures do not become task failures, while method-level operational incompatibility is included in unconditional method success according to the frozen failure policy. Gold obligations may be used only for external evaluation. The held-out tasks, their evaluator results, and any test failure are forbidden inputs to artifact generation or repair.

\paragraph{Artifact-to-result traceability.}
The public source package contains the nine full ACCs and their automatic and gold obligation files; 45 development and 225 held-out tasks; deterministic environments and evaluators; all 207 compressed artifacts and metadata; the 15,525-record terminal ledger; primary and secondary analysis outputs; executable analysis code; and the design, runtime, artifact, execution, and result-chain manifests. A formal artifact index maps each ACC--method--budget cell to its artifact and metadata hashes, and every ledger record carries that artifact identity. Package-level SHA-256 manifests bind the released files without exposing credentials, private endpoints, or local paths.

\paragraph{Additional validity boundaries.}
The budget grid is coarse, especially between 75\% and 50\%, so the frontier is an empirical curve rather than an estimated mathematical change point. Full-context success is 93.8\%, and compressed outcomes are reported absolutely rather than interpreted as harm only when the paired Full run succeeds. LLMLingua is represented by one frozen package/checkpoint configuration without manual repair. The study fixes a bare-JSON action protocol in deterministic sandboxes; other languages, native tool calling, live tools, schema-delivered documentation, and multi-agent communication remain outside its scope.

\end{document}